# Meta-Evaluation of Comparability Metrics Using Parallel Corpora

Bogdan Babych[1], Anthony Hartley[1]

[1] Centre for Translation Studies, University of Leeds, Woodhouse Lane,
Leeds LS16 5PT, UK
{b.babych, a.hartley}@leeds.ac.uk

**Abstract.** Metrics for measuring the comparability of corpora or texts need to be developed and evaluated systematically. Applications based on a corpus, such as training Statistical MT systems in specialised narrow domains, require finding a reasonable balance between the size of the corpus and its consistency, with controlled and benchmarked levels of comparability for any newly added sections. In this article we propose a method that can meta-evaluate comparability metrics by calculating monolingual comparability scores separately on the "source" and "target" sides of parallel corpora. The range of scores on the source side is then correlated (using Pearson's r coefficient) with the range of "target" scores; the higher the correlation – the more reliable is the metric. The intuition is that a good metric should yield the same distance between different domains in different languages. Our method gives consistent results for the same metrics on different data sets, which indicates that it is reliable and can be used for metric comparison or for optimising settings of parametrised metrics.

## 1 Introduction

In several areas of computational linguistics there is a growing interest in measuring the degree of 'similarity', or 'comparability', between different corpora or between individual texts within these corpora. Interpretation of the concept of comparability varies according to the intended application, but many areas share the idea that it is useful to have an automated metric which ranks corpora, sub-corpora or documents according to the degree of their 'closeness' to each other. Typically, closeness is either measured by pre-defined formal parameters (such as lexical overlap) or intuitively described in terms of less formal linguistic categories (such as *genre* or *subject domain*). In the present study, formal metrics are based on combinations of measurable parameters that correlate with human intuitions about the intended linguistic categories.

The concept of comparability is relevant both in the monolingual context (as similarity between corpora/texts written in the same language) and in the cross-lingual context (as similarity of corpora/texts in different languages). Later we give examples of the areas and applications where measuring corpus and text comparability is useful.

In the monolingual context the concept of corpus comparability is used in computational lexicography for building translation dictionaries (e.g., Teubert, 1996 [1]), and in corpus linguistics for identifying qualitative differences between language varieties (e.g., British vs. American English), domains, modalities (spoken vs. written language), in order, for example, to determine which words are particularly characteristic of a corpus or text' (Kilgarriff, 2001:233 [2]; Rayson & Garside, 2000 [3]). Another monolingual application is automatic identification of domains and genres for texts on the web (e.g., Kessler et al., 1998 [4]; Sharoff, 2007 [5]; Vidulin et al., 2007 [6]; Kanaris & Stamatatos, 2009 [7]; Wu et al., 2010 [8]), with the goal of developing domain-sensitive and genre-enabled Information Retrieval (IR) methods, which can restrict search according to automatically identified fine-grained text types (such as blogs, forum discussions, editorials, analytical articles, news, user manuals, etc.).

Cross-lingual comparable corpora are frequently used for identifying potential translation equivalents for words, phrases or terminological expressions (Rapp, 1995 [9]; Rapp 1999 [10]; Fung, 1998 [11]; Fung & Yee, 1998 [12]; Daille & Morin, 2005 [13]; Morin et al., 2007 [14]), or supporting human translators in dealing with non-trivial translation problems (Sharoff et al., 2006 [15]; Babych et al., 2007 [16]). Multilingual comparable corpora are now becoming increasingly useful in training translation models for Statistical Machine Translation (SMT): (Wu & Fung, 2005 [17]; Munteanu et al, 2004 [18]; Munteanu & Marcu, 2006 [19]), especially for under-resourced languages, where traditional parallel resources are not available, or are very small or in any other way unrepresentative (Vasiljevs, 2010 [20]). There are several dimensions of comparability, which can be summarised as follows:

(1) *granularity of comparability*: a measure of correspondence between units at different structural levels:
- corpus-level comparability – between corpus A and corpus B as a whole, or comparability between individual sections (subcorpora) within the corpus;
- document-level comparability – between different documents within or across corpora, e.g., (Lee et al, 2005 [21])
- paragraph- and sentence-level comparability – between structural and communicative units within or across individual documents, e.g., (Li et al., 2006 [22])
- comparability of sub-sentential units – between clauses, phrases, multiword expressions, lexico-grammatical constructions

(2) *degrees of comparability*: a level of closeness between two units of comparison on the scale ranging between close, then free translations (or plagiarised sections in the monolingual context), then texts about the same event, texts on the same topic, corpora in the same domain, and finally to completely unrelated corpora. In the cross-lingual context we can distinguish the following broad categories:

- parallel corpora (consist of translated documents, where alignment at the sentence level is possible, e.g., corpora collected from multilingual news websites)
- strongly comparable corpora (consist of texts describing the same event or subject, where alignment at the document level is still possible, e.g., linked Wikipedia articles in different languages, news stories on the same event)
- weakly comparable corpora (consist of texts in the same domain or genre, but describing different events or areas; document alignment is usually not possible, e.g., collection of British and German laws on immigration policy).

Specific applications of comparable corpora use a different understanding and different ways of identifying the intended degree of closeness between corpora or texts. However, in many cases these metrics use similar sets of features and similar methods of calculating the scores. For example, both monolingual and cross-lingual comparability in terms of subject domains typically rely on lexical features weighted or filtered by frequency, textual salience of key terms, etc. Often the only difference is that in the case of cross-lingual metrics lexical features (words) are mapped to words in another language using bilingual dictionaries or Machine Translation (MT) systems, while in monolingual applications lexical features are matched directly. These similarities mean that measuring comparability increasingly becomes a core technological challenge in Natural Language Processing that needs to be developed and evaluated systematically.

Many applications now require not just *it-looks-good-to-me* comparable corpora, but corpora with controlled and benchmarked levels of comparability according to certain criteria. Comparability metrics are used not only for reporting scores of closeness between corpora, but also ─for collecting additional texts to make a corpus bigger, or to filter out unwanted texts from corpora to ensure the intended level of comparability. From this perspective it is important to understand how reliable a particular metric is and to what extent it matches its specifications in its ability to *evaluate* comparability of corpora or individual texts. To date, we are not aware of any systematic research on such *meta-evaluation* (or calibration) of comparability metrics.

In this article we give an example of an application domain which potentially requires corpora with controlled levels of comparability. We propose a method that can meta-evaluate different metrics used to measure comparability. We show that our method gives consistent results for the same metrics on different data sets, which indicates that it is reliable and can be used for selecting a best-performing metric, or for finding the most efficient settings of parameters for parametrised metrics.

The rest of the article is organised as follows. In Section 2 we describe our application area: creating a coherent selection of parallel texts for training domain-specific MT systems. In Section 3 – Methodology – we describe different metrics that we use for measuring corpus-level comparability and our methodology for the meta-evaluation of these metrics. In Section 4 we present the results of this meta-evaluation, and in Section 5 we discuss conclusions and future work.

## 2 Application of corpus comparability: Selecting coherent parallel corpora for domain-specific MT training

Traditionally statistical and example-based MT have relied on parallel corpora (collections of texts translated by human translators) to train statistical translation models and automatically extract equivalents. However, a serious limitation of this approach is that translation quality is impaired where parallel resources are not available in sufficient volume.

Firstly, it has been shown that on average improving translation quality at a constant rate requires an exponential increase in the training data (e.g., Och and Ney, 2003: 43) [23], i.e., if improving some MT evaluation score, e.g., BLEU, by one point required doubling the size of a training corpus, then further improvement by one additional point would require a corpus four times bigger than the initial size, etc. This dependency imposes fundamental limitations on translation quality even for well-resourced languages, such as English, German or French, where only the huge data sets used by engines like Google Translate produce relatively good quality (and even then, only for certain text types). Smaller and less resourced languages do not have the benefit of such data repositories, which results in a much lower MT quality.

Secondly, training translation models and language models for SMT has been shown to be domain-dependent to a much greater degree than rule-based MT (RBMT) (Eisele, 2008: 181) [24]. If an SMT engine is trained on a corpus that doesn't match the domain of the translated text, then the quality for such out-of-domain translation becomes much lower. In practice this means that for more narrow subject domains and text types SMT cannot produce acceptable translation quality without domain adaptation, which needs correspondingly highly-specific parallel textual resources.

For translation to and from under-resourced languages in narrow domains and for specific text types the two problems described above are combined. As a result traditional ways of building SMT engines with acceptable translation quality are often not possible for many domain / language combinations.

There is, therefore, a need to develop a fine-grained monolingual domain selection and domain control mechanism for evaluating comparability of corpus sections that can usefully be added to any SMT training corpus (comparability here is measured monolingually – either on the source or on the target side). The methodology should allow MT developers to balance the size of the corpus to be built and its internal consistency, in terms of how newly added sections match its originally intended subject domain.

## 3 Methodology

The methodology for computing the comparability of sections for an MT training corpus is typically based on calculating the degree of overlap between the two files in terms of simple word tokens, or at more advanced levels of linguistic annotations, such as lemmas (dictionary forms of words), combination of lemmas and part-of-speech codes, translation probabilities for each of the words, etc. Note that there are several major challenges for the efficient calculations of this overlap.

Firstly, calculated scores for comparability should be consistent with human intuition about closeness between the two sections, and what constitutes the subject domain at different levels of granularity, e.g., the broader domain of computer hardware vs. a more narrow domain of network technologies, documentation for different types of network cards, etc. This is required if user needs for finer- or coarser-grained domains are to be adequately addressed for most types of projects.

Secondly, for practical applications the number of calculations between compared sections can be very large; so the calculation method should be fast enough to produce the results in real time.

Thirdly, comparison often needs to be done between sections of corpora file of different sizes, so the calculation method should be minimally affected by the size of the compared sections or texts.

Ideally, comparison should also take into account both source and target parts of new additions to corpora, and evaluate not only monolingual distance, but also the "translation" distance (which could mean that the same translation equivalents are used, and that terminology is translated consistently across the selected uploads).

## 3.1. Description of calculation method

The method which we use in our experiments is based on the work of Kilgarriff (2001) [2]. This method was initially developed for the purposes of linguistic analysis, i.e., to find words which that are substantially different in two corpora, e.g., a corpus of spoken vs. a corpus of written English. But one of the side-effects of this method is that it can produce a single numeric value that shows the 'distance' between the two compared monolingual corpora.

So in our work we focus not on identifying individual words which are used differently in different corpora, but on general quantitative measures of comparability between them. The method can be summarised as follows.

For each corpus (on the source or target side) we build a frequency list, and take the top 500 most frequent words (which also include function words).

Since corpora can be of different sizes, we use relative frequency (the absolute frequency, i.e., the number of times each word is found, divided by the total number of words in the corpus).

We compare corpora pairwise using a standard Chi-Square distance measure:

$$Chi^2 = \sum_{w_1}^{w_{500}} \frac{Freq(corpus^A) - Freq(corpus^B)}{Freq(corpus^A)}$$

## 3.2. Symmetric vs. asymmetric calculation of distance

The challenge for this method is that some words which are in the top-500 list for CorpusA may be missing from top-500 in CorpusB and vice versa. If the algorithm encounters the missing word, then it just adds its relative frequency to the value of the Chi-Square distance.

Obviously, exactly the same number of words is missing from the top-500 in CorpusA and in CorpusB. However, the sum of relative frequencies for these words can be different, e.g., it is possible that on average more frequent words will be missing from CorpusA, and less frequent from CorpusB.

Therefore, if we compute the Chi-Square distance from CorpusA to CorpusB, and then from CorpusB to CorpusA, we will get different values, which shows that the term 'distance' (used in an everyday sense) is not exactly right for describing the values: our calculation method is asymmetric.

Instead, Kilgarriff (2001) [2] uses a symmetric calculation: he takes into account only words which are common in both corpora, and goes down the frequency lists as far as it is needed to collect the 500 most frequent common words. This method always returns the same values of distance for any direction. However, the symmetric approach has its drawbacks: missing words do not contribute to the score directly (only by virtue of occupying 'someone else's place'); also it is harder to select the initial list for comparison: in the worst case scenario it is necessary to start with top-1000 words for each of the corpora, and then to remove mismatches. It may take slightly longer to do these calculations, and in real time this may result in unnecessary delays and increased waiting time for users.

In our approach we make two independent asymmetric calculations in both directions: CorpusA →CorpusB and CorpusB →CorpusA, and get two Chi-Square scores.

However, now is not obvious what is the best way to combine these two scores into a single measure of distances between the corpora: one method would be to take the Average of the distances; another method is to take the Minimum distance as the value.

We experimentally compare these two possibilities in later sections, and show that the minimum of the two Chi-Square scores computed for the two directions gives better and more meaningful results.

## 3.3. Calibrating the distance metric

We used corpora available from TAUS (Translation Automation User Society) in its TDA (TAUS Data Association) repository. Corpora there were initially annotated by data providers in terms of 'subject domains', which are identified manually at the upload stage. The idea is that the metric should be able to simulate identification of these domains automatically.

We calculated the distance between different sections of TDA repository – individual uploads and collections of uploads grouped by the same data provider and domain. In order to tell whether the metric intuitively makes sense, we checked whether there is an agreement between the resulting values and the labels provided by the TDA members.

In our experiment we focussed on the English (US and UK) – French (France) language pair. We selected the set of uploads in a way which covered different combinations of domains and data providers: some corpora are labelled as belonging to different domain, but were produced by the same company. Some were produced by different companies but were labelled with the same domain tag.

These labels were used as a benchmark for judging the quality of the lexical comparability metric. We aimed at giving the smallest distance score to corpora within the same subject domain.

The results of measuring comparability between sections of the corpus given by different data providers are presented in Figure 1. Different shades of grey visualise different ranges of distances: the closer the distance, the darker the colour.

| | compHardB | compHardA | compHardC | compSoftJ | compSoftK | compSoftC | compSoftG | compSoftF | consumerElectronicsD | financialE | legalEC | hansardEuroparl | legalUN | pharmBiotechH | profBusinessK | profBusinessG | profBusinessD |
|---|---|---|---|---|---|---|---|---|---|---|---|---|---|---|---|---|---|
| compHardB | | 0.18 | 0.21 | 0.42 | 0.47 | 0.5 | 0.41 | 0.28 | 0.24 | 0.26 | 0.44 | 0.8 | 0.91 | 0.38 | 0.51 | 0.43 | 0.25 |
| compHardA | 0.18 | | 0.33 | 0.4 | 0.52 | 0.49 | 0.49 | 0.27 | 0.29 | 0.27 | 0.5 | 0.65 | 0.77 | 0.45 | 0.54 | 0.41 | 0.29 |
| compHardC | 0.21 | 0.33 | | 0.45 | 0.44 | 0.5 | 0.48 | 0.32 | 0.35 | 0.37 | 0.44 | 0.74 | 1.01 | 0.45 | 0.47 | 0.55 | 0.36 |
| compSoftJ | 0.42 | 0.4 | 0.45 | | 0.6 | 0.64 | 0.53 | 0.39 | 0.46 | 0.46 | 0.53 | 0.64 | 0.83 | 0.46 | 0.67 | 0.59 | 0.47 |
| compSoftK | 0.47 | 0.52 | 0.44 | 0.6 | | 0.81 | 0.43 | 0.56 | 0.49 | 0.48 | 0.85 | 0.82 | 1.5 | 0.63 | 0.03 | 0.68 | 0.5 |
| compSoftC | 0.5 | 0.49 | 0.5 | 0.64 | 0.81 | | 0.64 | 0.54 | 0.51 | 0.56 | 0.68 | 0.83 | 1.03 | 0.44 | 0.64 | 0.73 | 0.52 |
| compSoftG | 0.41 | 0.49 | 0.48 | 0.53 | 0.43 | 0.64 | | 0.44 | 0.47 | 0.52 | 0.56 | 0.9 | 1.12 | 0.55 | 0.46 | 0.79 | 0.48 |
| compSoftF | 0.28 | 0.27 | 0.32 | 0.39 | 0.56 | 0.54 | 0.44 | | 0.37 | 0.41 | 0.56 | 0.58 | 0.95 | 0.48 | 0.62 | 0.66 | 0.38 |
| consumerElectronicsD | 0.24 | 0.29 | 0.35 | 0.46 | 0.49 | 0.51 | 0.47 | 0.37 | | 0.27 | 0.44 | 0.72 | 0.94 | 0.38 | 0.53 | 0.38 | 0.01 |
| financialE | 0.26 | 0.27 | 0.37 | 0.46 | 0.48 | 0.56 | 0.52 | 0.41 | 0.27 | | 0.44 | 0.67 | 0.79 | 0.4 | 0.51 | 0.35 | 0.27 |
| legalEC | 0.44 | 0.5 | 0.44 | 0.53 | 0.85 | 0.68 | 0.56 | 0.56 | 0.44 | 0.44 | | 0.74 | 0.73 | 0.43 | 0.93 | 0.7 | 0.45 |
| hansardEuroparl | 0.8 | 0.65 | 0.74 | 0.64 | 0.82 | 0.83 | 0.9 | 0.58 | 0.72 | 0.67 | 0.74 | | 0.87 | 0.6 | 0.92 | 0.84 | 0.72 |
| legalUN | 0.91 | 0.77 | 1.01 | 0.83 | 1.5 | 1.03 | 1.12 | 0.95 | 0.94 | 0.79 | 0.73 | 0.87 | | 0.88 | 1.48 | 0.8 | 0.91 |
| pharmBiotechH | 0.38 | 0.45 | 0.45 | 0.46 | 0.63 | 0.44 | 0.55 | 0.48 | 0.38 | 0.4 | 0.43 | 0.6 | 0.88 | | 0.68 | 0.66 | 0.38 |
| profBusinessK | 0.51 | 0.54 | 0.47 | 0.67 | 0.03 | 0.64 | 0.46 | 0.62 | 0.53 | 0.51 | 0.93 | 0.92 | 1.48 | 0.68 | | 0.72 | 0.53 |
| profBusinessG | 0.43 | 0.41 | 0.55 | 0.59 | 0.68 | 0.73 | 0.79 | 0.66 | 0.38 | 0.35 | 0.7 | 0.84 | 0.8 | 0.66 | 0.72 | | 0.38 |
| profBusinessD | 0.25 | 0.29 | 0.36 | 0.47 | 0.5 | 0.52 | 0.48 | 0.38 | 0.01 | 0.27 | 0.45 | 0.72 | 0.91 | 0.38 | 0.53 | 0.38 | |

Figure 1. Chi-Square distances between different data providers (labels indicate domain and owner, e.g., compSoftG is the label 'computer software' produced by the company G)

It can be seen from Figure 1 that the metric reliably identifies the following:

(1) all corpora within the 'computer hardware' domain are reliably grouped together, and distinguished from other domains;

(2) some of the corpora which were produced by the same companies 'D' and 'K' were reliably grouped together, even though the corpora had received different human labels: company D - 'consumer electronics' and 'professional business services'; company K - 'computer software' and 'professional business services'. These instances can be explained by inconsistency in assigning labels to corpora which essentially represented the same domain.

(3) different domains which are intuitively close to each other were also grouped together: 'computer hardware' and 'consumer electronics', and then at some distance – several corpora on computer software.

However, there are several problems with the presented distances and grouping:

(1) the 'computer software' and 'legal' domains are not coherently grouped. A possible reason is a greater variety of sub-domains within the 'computer software' domain (it may describe more 'products', and have more diverse lexical profiles);

(2) the 'pharmaceutical and biotechnology' and 'financial' domains are not sufficiently distinct from the 'software' and 'hardware' domains.

Still these problems can be attributed to inconsistencies in human labelling, as well as to shortcomings of the metric itself. Symbolic labels are speculative in their nature, and do not capture the inner structure or diversity of the domain; at present human annotation offers no way of dealing with mislabelled data.

## 4. Validation of the scores: cross-language agreement for source vs. target sides of TMX files

We validate our choice of metric by comparing different versions of Kilgarriff's metric for computing the distance between corpora. As we indicated, there are two possibilities for combining asymmetric Chi-Square distances: we can either take the average of the two different values, or go for the minimum of the two values.

Sections of corpora in the TDA repository are uploaded in TMX (Translation Memory Exchange) format, which is an XML file with sentence-or segment-aligned parallel corpora.

The idea for comparison is the following: we use each of the possibilities on the source side and on the target side of the same TMX files and then compare how the scores "agree" with each other. The agreement can be measured by standard statistical method for calculating correlation, like Pearson's r correlation coefficient: if there is a good agreement, r is closer to 1 or to -1; if there is no agreement r is closer to 0.

To get more data points for more reliable calculation of correlation we further split sections of the corpora presented in Figure 1 into individual uploads, e.g. for Hardware Company A we had 5 individual TMX files. Distances were computed at this finer granularity between all individual uploads.

If one of the compared metrics produces a higher correlation, then it means that the results obtained on the source side are more consistent with the results obtained on the target side, and the metric is more meaningful. Essentially, we know from the start that the two texts came from the same TMX, but the metric doesn't have that information. The better it can figure this out, the more reliable it is.

Table 1 compares the r correlation figures for individual uploads. It can be seeing from the table that the minimum distance has the best correlation between source and target sides of TMX: r=0.85, and can be viewed as a more reliable metric compared to the average distance or the third score we computed, the one-direction distance.

| Metric | r-correlation |
|---|---|
| Minimum distance | **0.85** |
| Average distance | 0.67 |
| One-direction distance (A->B and B->A) | 0.61 |

Table 1. Pearson's r correlation for distances computed for English vs French

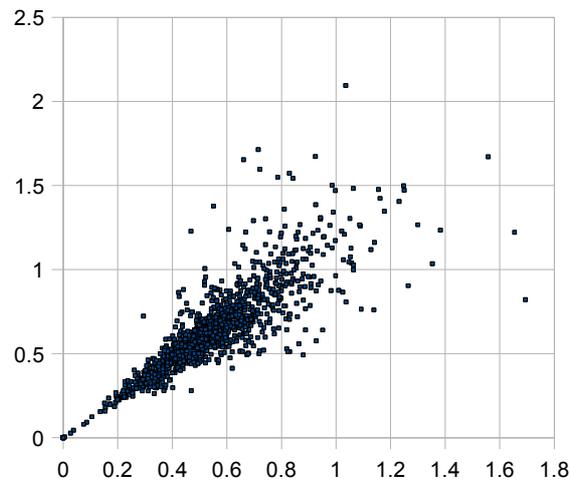

Figure 2. Minimum Chi-Square distance (x=En; y=Fr)

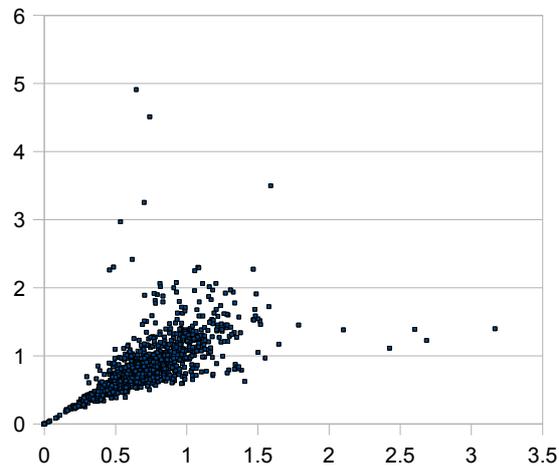

Figure 3. Average Chi-Square distance (x=En; y=Fr)

Figure 2 and Figure 3 further illustrate this difference: they compare the correspondence between the TMX distances for source text (horizontal axis) vs. distances for target texts for the same uploads (and illustrate the correlation figures presented above. Figure 2 indicates the distances in terms of minimum chi-square scores for TMX-A →TMX-B vs. TMX-B →TMX-A. Figure 3 indicates average chi-square scores for the same pairs of distances.

It can be seen that the minimum distances have a much better correlation between source and target, so they more reliably indicate whether the texts are indeed closer to each other.

This method offers a way to evaluate different comparability scores: the more the source and target agree with each other, the better the quality of the matching scores is. This evaluation method is based on the assumption that if the texts are close in terms of the source side, the scores should also show that they are close in terms of the target side.

However, there is a question of how to interpret divergences of the dots from the diagonal line. One explanation is that the quality of the matching scores is not so good. But another explanation suggests that some inconsistent translation equivalents are used across the upload in the target, so even if the documents are genuinely close on the source side, they become divergent on the target side. These issues require a more careful look into the compared data.

To verify that our meta-evaluation method provides consistent results for different sizes of evaluated corpus we repeated the experiment for the same language pair, but now we used the original joined TMX files, where all uploads are grouped together for the same data provider. Figure 3 shows an agreement for minimum chi-square scores for TMX-A →TMX-B vs. TMX-B →TMX-A. Correlation coefficients for this setting are presented in Table 2.

The highest correlation is again found between minimum scores across the two directions, which confirms our choice of this metric as the most reliable.

| Metric | r-correlation |
|---|---|
| Minimum distance | **0.88** |
| One-direction distance (A->B) | 0.72 |
| One-direction distance (B->A) | 0.86 |

Table 2. Pearson's r correlation for English vs French on larger corpus sections

These results shows that data with a different number of data points obtained on sections of different sizes point to the same metric as the best one, which indicates that our proposed method is internally coherent.

## 5. Discussion and future work

We have proposed a method for meta-evaluation of comparability metrics using correlation between source and target sides of parallel corpora. We used a collection of parallel corpora available from the TDA repository. Comparability metrics need to be calibrated on a diverse parallel corpus that includes sections with several distinct and annotated subject domains, genres, etc. However, after successful calibration such comparability metrics can be further used to collect monolingual and bilingual comparable corpora, without the need to have expensive parallel resources. Pearson's r coefficient, which we calculate during calibration, gives an indication of how reliable the metric is and how much noise might occur in the data.

The metric which was found to perform best in our experiment (minimum Chi-Square distance between the compared top-500 words of frequency lists) has relatively high agreement for data generated on the source and target sides of TMX files (r=0.85), which indicates the upper limit of the metric's reliability.

However, applicability of the proposed meta-evaluation method is limited by the accuracy and completeness of translations in the parallel corpus used for calibration: gaps or excessively free translation can result in shifts in feature patterns, so distances between different domains calculated on source and target texts can become greater. Another potential limitation of the method is its reliance solely on those formal parameters which can be computed in a language-independent way, and do not vary substantially across languages. Language-specific differences, e.g., variations in type-token ratio (due to different morphological structure of languages) can potentially lead to differences in feature patterns and, as a result, lower correlation figures.

In future work we will investigate to what extent our method is limited by the quality of the translated parallel corpora and by language-specific features – by measuring the comparability of real corpora. We will also explore ways to externally validate the proposed method.